# Novel Design and Implementation of a Vehicle Controlling and Tracking System

Hasan K. Naji[1], Iuliana Marin[2], Nicolae Goga[3], Cristian Taslitschi[4]
University "POLITEHNICA" of Bucharest, Romania
Doctoral School Automatic Control and Computers
Doctor of Philosophy in Computers and Information Technology

*Abstract*—The purpose of this project is to build a system that will quickly track the location of a stolen vehicle, thereby reducing the cost and effort of police. Moreover, the vehicle's computer system can be controlled remotely by the owners of the vehicle or police. More precisely, the goal of this work is to design a, develop remote control of the vehicle, and find the locations with Latitude (LAT) and Longitude (LONG).

## I. INTRODUCTION

Vehicle theft is an existing phenomenon which is considered a common occurrence that extends beyond its security dimensions to economic, social, and psychological damages. Despite all the recent techniques and actions that have been taken to avoid vehicle theft, the National Crime Information Center (NICC) stated at the end of December 2014 that millions of vehicles were stolen and billions of dollars lost [1]. The same is true six years later. Therefore, many researchers address this crucial requirement to protect lives and assets.

The key concept of this research to design a system using the Global Positioning System (GPS), the Global System for Mobile communications (GSM) and a microcontroller that can control the vehicle when it is stolen [2,3,4,5]. GSM is directly connected to the microcontroller (which is an Arduino processor) and then connect the entire system to the GPS. The Short Message Service (SMS) is processed in the microcontroller unit and forward it to the GPS unit to give the exact location in the form of latitude and longitude on the owner's mobile phone. Some of the features of the system are: the ability to lock and unlock vehicle doors, find its position, turning on the lights, turn flashing lights [6].

Through this work, simple techniques used by everyone will be used to work on solving the problem of tracking the vehicles if they are stolen. The use of messaging technology in mobile phone networks is the cheapest, easiest and best way to communicate between the vehicle and its owner, through which the type of security of the system was determined by creating messages for each vehicle owner as desired. However, we achieved customer requirements as desired, by specifying the type of numerical messages or by letters. In what follows we present a literature review. After that we describe the system and at the end we draw the conclusions.

## II. LITERATURE REVIEW

In the research of Rani et al. [7], it is described a system that is based on biometrics framework used for securing a vehicle. Biometric technology in the form of the driver's fingerprint recognition is used to start the engine. The system operates in two modes. The first mode is "*conformity*", which controls the proper operation of the vehicle, while the second mode is "*nonconformity*", which will alert the vehicle owner automatically by sending a message with GSM. It is also possible for this system not to allow a drunk driver or a driver when s/he fall asleep.

Sugumaran et al. presented a monitor system based on a "Raspberry pi" and an infrared sensor [8]. An infrared sensor is used to identify a specific area when many people enter an area. The system will turn on automatically when the system senses a movement. Then the information is sent to the owner's mobile.

A system designed for controlling a vehicle is also described by Kumar et al. [9]. This is done by controlling the fuel injection using an electronic solenoid valve through which a microcontroller controls the driver circuit. A passkey will be given to the owner of the vehicle in case of conformity and the solenoid valve will open and the vehicle will start. In case of unconformity the program will send to the owner a message through a 'GSM' modem connecting an alarm.

In the research of Ramadan et al. [10], it is presented an anti-theft and tracking system for the vehicles. By the direct connection with the owner of the vehicle the system can detect the areas and current settings of the vehicle, as well as, detecting the locations of a group of vehicles. This is done by using the "Google Earth".

A system created for mobile vehicle security provides a connection between the driver and the system, as described in the research of Shah et al. and Abdullah [11, 12]. In the case of intrusion, the system will send a '*warning messag*e' to the owner. Once the message is sent to the vehicle owner, she/he will be able to control all the vehicle safety features through his smartphone.

In the paper of Pany et al. [13], it is described a system installed in the engine of a vehicle with a 'GSM' modem connected to a microcontroller. The system offers multiple features to the owner. One such feature includes a password-protected engine start. If the password is correct, the system will authenticate and the car will start working. In case of non-conformity, the engine will not start and the siren will go on and the system will send a message to the car owner through the 'GSM' system [13].

As compared with the previous described system, ours has several distinctive features. For example, the system relies on two types of energy sources: the car battery and the portable battery that is used in case of an event of a car battery failure. The system allows the capability to replace the phone number for the system and the owner. Several operations were created on the vehicle that can be extended into multiple ones to control all car parts. The system has a high degree of security through the confidentiality of phone numbers and the way to verify them through the code which is used by the Arduino processor. If the authentication process succeeds, the system will begin to execute orders coming from the owner.

## III. RESEARCH METHODOLOGY

The design is made in such a way to create an efficient process for real-time tracking and controlling any vehicle equipped with this technology, anywhere at any time. Our system which uses a microcontroller and traditional smartphone technologies can be done at lower cost and with larger flexibility as compared to other systems. The system works to integrate some global systems which are reliable and are able to give real results to the user, such as GPS, GSM and smartphone technology which are the most used methods to control and track vehicles [14,15,16,17,18]. The process of sending messages and executing the request is impacted by some problems, the most important one being the signal strength of the system network. The signal strength will affect the performance of the current system technology. It typically takes this system from four to six seconds to interact with the message and implementation.

The system can be developed significantly and efficiently to suit the evolving electronic world based on two directions. The first direction is to control the vehicle, while the second part is to easily track its location. This project uses an Arduino microcontroller to connect various peripheral devices [19, 20, 21, 22, 23, 24,25, 26]. The system monitors constantly the vehicle and determine the local area of the vehicle upon request in real-time. Many control and tracking systems were designed to help persons or companies with a large number of vehicles. Those systems can manage and reduce the cost and effort of the police staff, in the case that a vehicle was stolen, within a short period of time. What differentiates this system is real-time tracking and feedback.

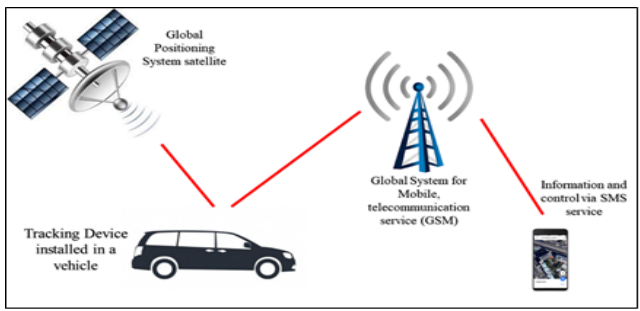

Fig 1. System Idea.

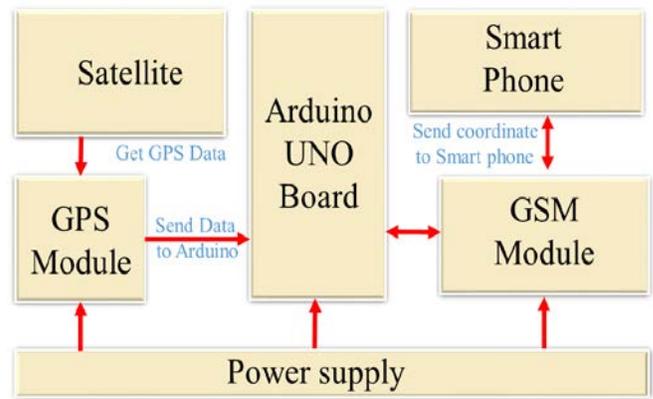

Fig 2. System Architecture.

The use of SMS applications has become popular because it is affordable, effective, convenient and easy to access, for transmitting and receiving data with high reliability [27, 28]. Therefore, this technology will be used in the current system, as illustrated in Fig. 1.

The system relies on a GPS receiver, modem GSM, and the as depicted in Fig. 2. The users can control, monitor, and interact with the vehicle from a dedicated application, as well as via Google Maps. Consequently, they can perform several actions through the vehicle, if it was stolen.

The system needs verification of the cellular phone number of the incoming message, to make sure that it is the owner's phone number, because it will use the Subscriber Identity Modules (SIM) phone number to authenticate the authorized person. If someone mistakenly sends a short message SMS to this system, she/he cannot receive any reply [29]. It would also be possible to contact the owner of the vehicle through a GSM kit with a portable computer to track the vehicle in real-time using Google Maps.

## IV. SYSTEM IMPLEMENTATION, TESTING AND RESULTS

### A. Hardware Implementations

The system can be run through a PC via direct connection or by connecting batteries directly to the system, as shown in Fig. 3.

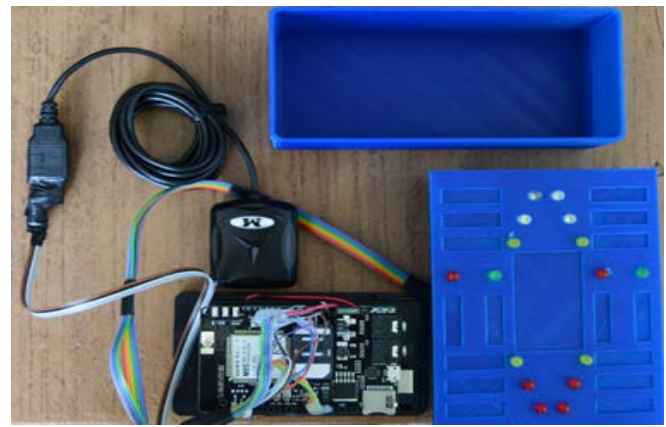

Fig 3. Finial Implementation of the System.

*B. Connect the GSM with Arduino*

At this step, the connection between the GSM and Arduino is made using the kit connector which is designed to connect between two different shields, as shown in Fig. 4.

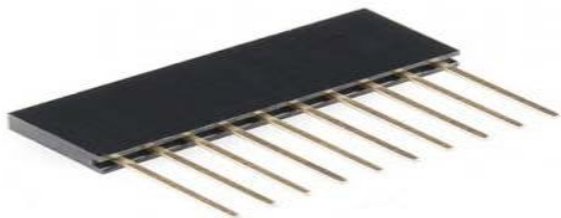

Fig 4. Kit Connector.

It must be fixed to the connector by the *"GSM"* to *"GPR Shield"*, which indicates each pin that specified in the table as shown below in four parts.

*C. Electrical Connection*

The digital connection alongside Arduino requires a corresponding pin for GSM, which is on the right side, as shown in Fig. 5. It can be noticed that there exists a connection between two shields, namely pin 7 in Arduino corresponds to D7 for GSM.

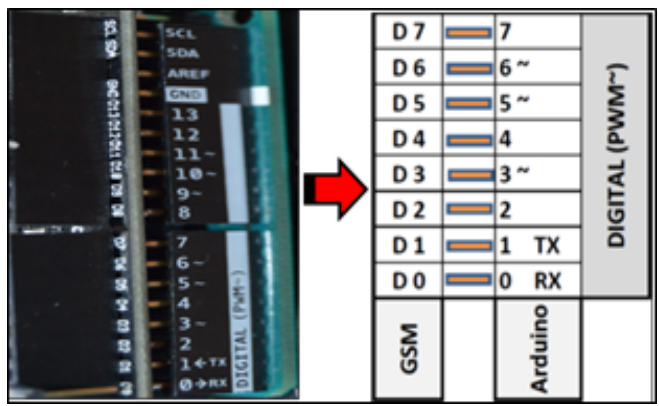

Fig 5. Connection between GSM and Arduino DIGITAL (PWM~).

*D. Arduino ANALOG IN*

The analog connection is alongside Arduino. Each pin of GSM is matched, as shown in Fig. 6. The connections between two shields are inversely related: for example, A5 pin in Arduino corresponds to A0 in the shield.

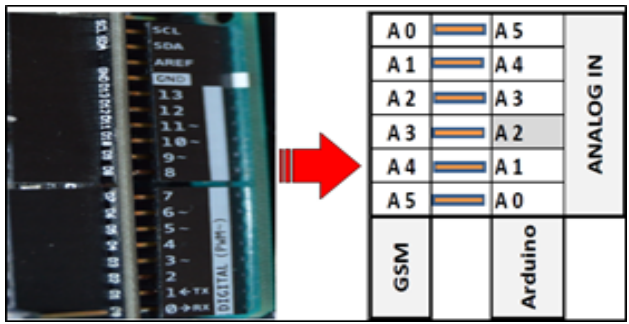

Fig 6. Connect GSM to the Arduino ANALOG IN.

*E. Power*

The system works by depending on the vehicle battery or the battery attached to it, the portable one, in the case if the vehicle's battery fails [3]. The power connection alongside with Arduino is required. Each pin of the GSM is matched as in Fig. 7, where the symmetry connections between two shields can be observed.

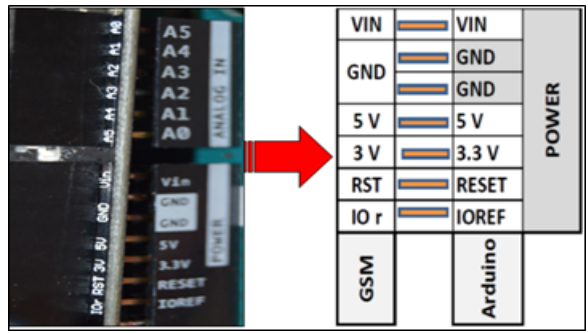

Fig 7. Connect GSM to the Arduino POWER.

*F. Board Connection with Multiplexer*

The circuit connection contains LEDs and a multiplexer to indicate all actions, as shown in Fig. 8. This board connects to Arduino along with cables.

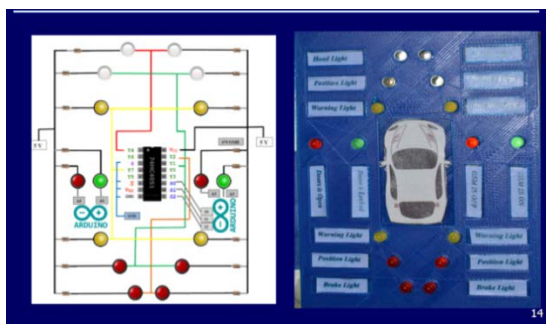

Fig 8. Circuit Diagram Control Box.

*G. Connection between GPS and Arduino*

The connection between GPS and Arduino is done using the steps 1 and 2 from Fig. 9, and through the RX and TX pins, corresponding to steps 3 and 4.

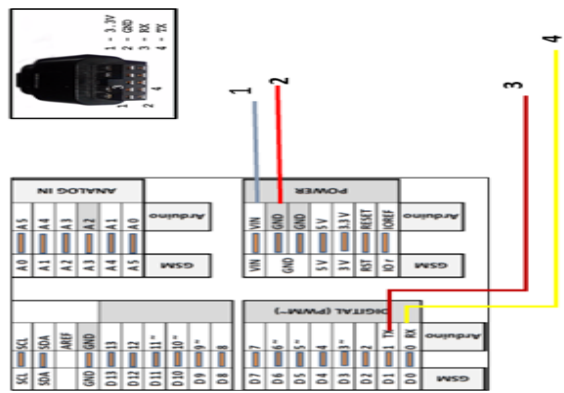

Fig 9. Connection between GPS and Arduino.

TABLE I. ACTION TABLE FOR THE VEHICLE SYSTEM

| Action | Stats | white | Red | Yellow | Green |
|---|---|---|---|---|---|
| Turn on the scene light of car | ON | 2 | 2 | | |
| Turn off the scene light of car | OFF | 2 | 2 | | |
| Turn on the front light of car | ON | 2 | | | |
| Turn off the front light of car | OFF | 2 | | | |
| Turn on the brake light of car | ON | | 2 | | |
| Turn off the brake light of car | OFF | | 2 | | |
| Turn on the flashing light of car | ON | | | 4 | |
| Turn off the flashing light of car | OFF | | | 4 | |
| Closed the doors of the car | ON | | | | 1 |
| Normal status the door is open | ON | | 1 | | |
| GSM is on after 60 second | ON | | | | 1 |
| GSM of off when the system is start | ON | | 1 | | |

The procedure listed in the Table I shown shows the LEDs indicators and correspondent action which is performed in the project.

## V. SOFTWARE IMPLEMENTATION

### A. Initialize

In the initialization procedure, first it is included the GSM and TinyGPS library; then it determines the GPS connection and SIM card number, and configures the library as shown in the figure. An array is defined to hold the number of "SMS" that will be received and shows the sender's number. Then a message from a phone is received - the message corresponds to an action that is going to be performed

### B. The Definition of Multiplexer Pin

The definition of S0, S1, and S2 along with 10, 11, and 12 to make the action that is needed on the board and specify the second action that works when the system starts (ON):

- The red *"LED"* - waiting; *"GSM"* is not active
- The Green *"LED"*: the "GSM" is active (wait approximately 30 seconds to be active).

Standby time allows *"GSM"* to figure out the network it operates on and the usual time duration for a transmission from the cell phone. The procedures run automatically when the system starts. The system works properly if the red *"LED"* for the '*GSM*' does not change to the green *"LED"*. This means that the system is an error and cannot send any message to the system:

- The red "LED" ---- the doors are opened
- The Green *"LED"* ---- the doors are locked

### C. Define the Commands

The programming commands for text messages that will be sent to the system must be specified, by making a matrix of letters for each message, for one particular action in the system. Herein, the user must be alerted to the nature and form of the message sent if it is in uppercase, lowercase letters, or even spaces between letters. Any text message (not identical to the messages installed in the system) will be ignored and therefore will not result in any apparent action, which gives the impression that the system does not work. Fig. 10 showing all the messages used in this system.

```
//define commands
char lights_on[]       = {'0','l','i','g','h','t','s',':','O','N'};
char lights_off[]      = {'1','l','i','g','h','t','s',':','O','F','F'};
char head_lights_on[]  = {'2','h','e','a','d',':','O','N'};
char head_lights_off[] = {'3','h','e','a','d',':','O','F','F'};
char brake_lights_on[] = {'4','b','r','a','k','e',':','O','N'};
char brake_lights_off[]= {'5','b','r','a','k','e',':','O','F','F'};
char warning_on[]      = {'6','w','a','r','n','i','n','g',':','O','N'};
char warning_off[]     = {'7','w','a','r','n','i','n','g',':','O','F','F'};
char location_on[]     = {'8','l','o','c','a','t','i','o','n',':','O','N'};
char location_off[]    = {'9','l','o','c','a','t','i','o','n',':','O','F','F'};
char doors_on[]        = {'+','d','o','o','r','s',':','O','N'};
char doors_off[]       = {'-','d','o','o','r','s',':','O','F','F'};
```

Fig 10. Definition the Commands.

### D. The Definition of the Commands Length

Each command has a fixed length as shown in the code bellow

```
//define vector length
int length_lights_on   = 9;
int length_lights_off  = 10;
int length_head_on     = 7;
int length_head_off    = 8;
int length_brake_on    = 8;
// etc.
```

### E. Initialize States of the System

The initial value of each feature and each *'LED'* in the initial state is OFF (in the case of a restart, the system will return all values in the initial state).

### F. Loop Function of the System

Table II we give the correspondence between messages, correspondent action and function performed by multiplexer.

TABLE II. MESSAGE TEXT, ACTION, FUNCTION

| Message Text | Actions | Function |
|---|---|---|
| 0lights: ON | Turn on position lights | Multiplex(1.0.0) |
| 1lights: OFF | Turn off position lights | Multiplex(1.0.1) |
| 2head: ON | Turn on head lights | Multiplex(0.0.1) |
| 3head: OFF | Turn off head lights | Multiplex(1.0.1) |
| 4brake: ON | Turn on brake lights | Multiplex(0.1.0) |
| 5brake: OFF | Turn off brake lights | Multiplex(1.0.1) |
| 6warning: ON | Turn on warning lights | Multiplex(1.1.1) |
| 7warning: OFF | Turn off warning lights | Multiplex(1.0.1) |

### G. Sending the Locations

"*Sending Location for the vehicle*" is a significant step in the project. Every moment the latitude and longitude of the system is sent such that we know where the vehicle is at any moment.

```
void get_locaton(bool newData2){
for (unsigned long start = millis();
millis() - start < 1000;)
{
while (Serial.available())
{
char c = Serial.read();
if (gps.encode(c));
newData2 = true;
}
}
if (newData2)
{
float flat, flon;
unsigned long age;
gps.f_get_position(&flat, &flon,
&age);
Serial.print("LAT=");
LAT = (flat ==
TinyGPS::GPS_INVALID_F_ANGLE ? 0.0 :
flat);
Serial.print(LAT, 6);
Serial.print(" LON=");
LON = (flon ==
TinyGPS::GPS_INVALID_F_ANGLE? 0.0:
flon);
Serial.print(LON, 6);
Serial.print(" SAT=");
SAT = gps.satellites() ==
TinyGPS::GPS_INVALID_SATELLITES ? 0:
gps.satellites();
Serial.print(SAT);
Serial.print(" PREC=");
Serial.println(gps.hdop() ==
TinyGPS::GPS_INVALID_HDOP ? 0 :
gps.hdop());
}
```

### H. Sending Google link

The most important part of this project is to send a Google link to show the location of the car at any moment in real time. This is done by first invoking the previous step to do that and then use the longitude and latitude to bring them to a required format. An example is given bellow: 'https://www.google/maps/place/44.44212+26.04938/@44.44212,26.04938,17z/data=!3m1!4b1! 4m2! 3m1! 1s0x0: 0x0? hl = en'.

By clicking on the link, we enable the Maps application to open in the smartphone to show the location for the vehicle. the code for sending the google link to SMS `//https://www.google.ro/maps/place/44.44212+26.04938/@44.44212,26.04938,17z/data=!3m1!4b1!4m2!3m1!1s0x0:0x0?hl=en`

```
if(send_google_maps == 1){
Serial.println("Acquire Location:");
get_locaton(newData);
//Composing the text for google the
message
txtMsg=
"https://www.google.ro/maps/place/";
dtostrf(LAT,6,6,LAT_buffer);
for(int i=0;i<=7;i++){
txtMsg= txtMsg + LAT_buffer[i];
}
txtMsg = txtMsg + "+";
dtostrf(LON,6,6,LON_buffer);
for(int i=0;i<=7;i++){
txtMsg= txtMsg + LON_buffer[i];
}
txtMsg = txtMsg + "/@";
for(int i=0;i<=7;i++){
txtMsg= txtMsg + LAT_buffer[i];
}
txtMsg = txtMsg + ",";
dtostrf(LON,6,6,LON_buffer);
for(int i=0;i<=7;i++){
txtMsg= txtMsg + LON_buffer[i];
}
txtMsg = txtMsg + ",17z/";
//Serial.println(txtMsg);
sendSMS();
send_google_maps = 0;
}
}
```

### I. Testing the System

At this point, every "SMS" sent to the system will be tested for all actions that are needed to control the vehicle. The figures from 11-16 show the types and forms of messages that are sent, as well as show all the actions in the system, such as, turn on/off warning light. Moreover, all the processing (send and receive) messages that occur in the Arduino.

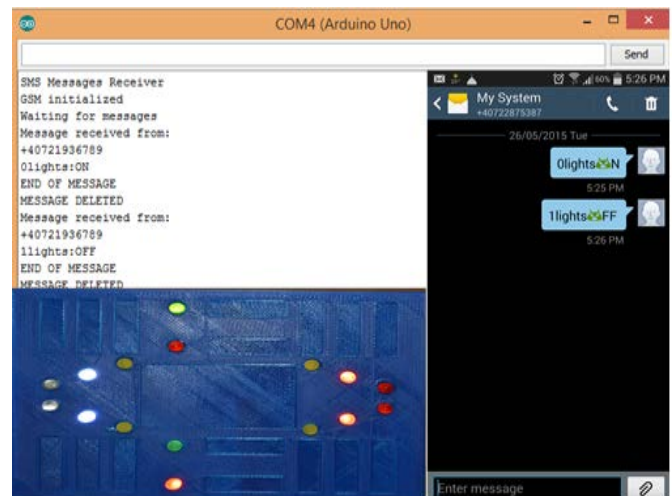

Turn on / off positions lights figure

Fig 11. Test the System ON/OFF Position Lights with Smartphone.

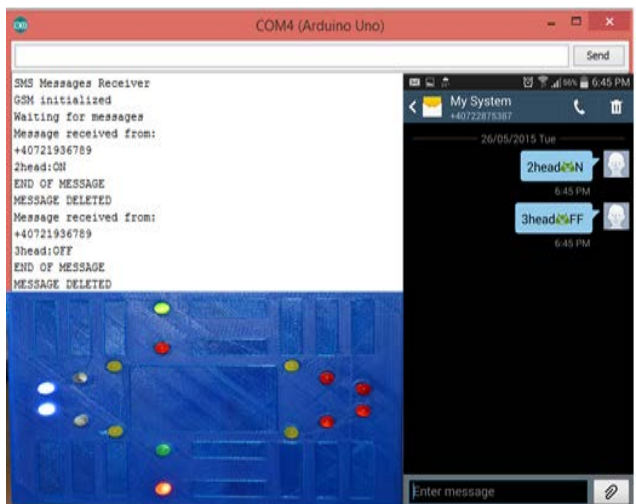

Turn on / off Headlights

Fig 12. Test the System ON / OFF Heads Lights with Smartphone.

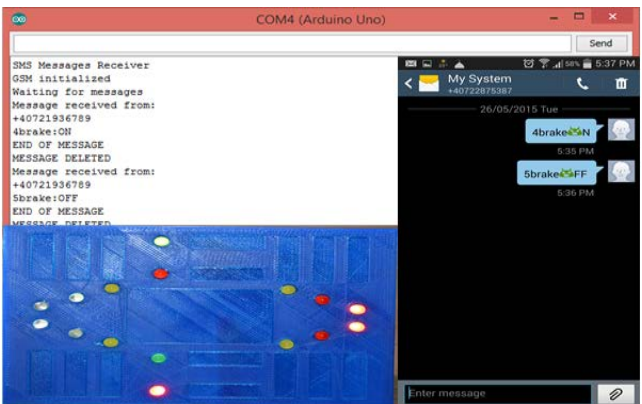

Turn on / off Brake lights

Fig 13. Test the System ON / OFF Brake Lights with Smartphone.

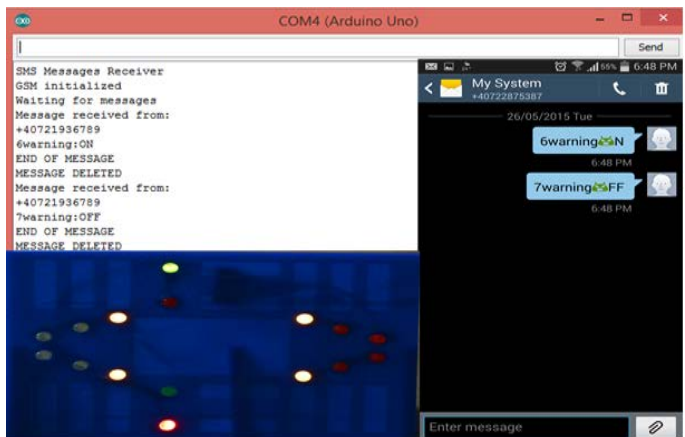

Turn on / off Warning lights

Fig 14. Test the System ON / OFF Warning Lights with Smartphone.

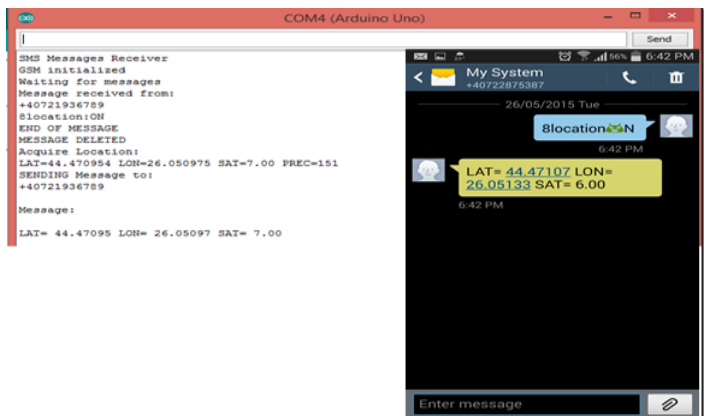

Send the Locations

Fig 15. Test the System Locations SMS with Smartphone.

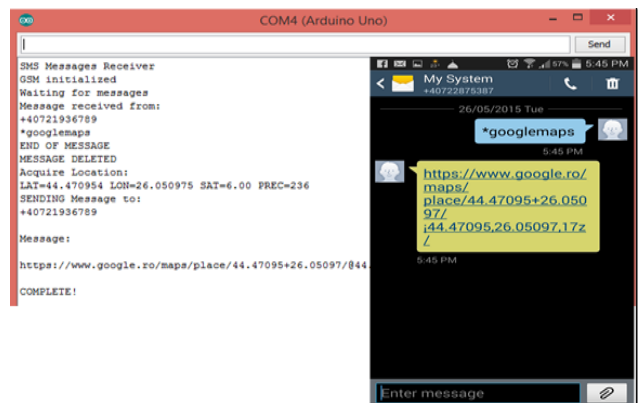

Send the Google Maps Link

Fig 16. Test the System Google Link with Smartphone.

## VI. THE RESULT

For the control of the vehicle through this system to be implemented on the actual destination inside the car, we need to add other devices are (Riley) to support operations to be achieved actually inside the car.

The process of sending the message and execution of the request interspersed with some problems, including signal strength of the network that are working on the system (mobile phone network) where ensure signal strength speed of implementation of the system from the moment of sending the message to the moment of execution where it normally take this system from 4 to 6 seconds to interact with the message and implementation (in case the message was written correctly).

Sending text messages is incorrect will not give any reaction by the system, which gives the impression that the system does not function properly for this reason had to have interest in the message, as noted in previous chapters in this regard.

## VII. CONCLUSION

The main objective of this system is to control of a vehicle and find out where it is in the case of theft. This system is made of several technologies including smartphone technology to find out the location area of the vehicle at a certain time. The system brings technology by using SMS text messages to reduce the costs resulting from the use of communication networks for mobile phones. It also uses several sources of power. This ensures system to work continuously in different conditions, even when the vehicle battery stops and gives reliability and independence to the system. Another important feature is the use of Google Maps.